\documentclass{article}

\usepackage{times}
\usepackage{bm}
\usepackage{graphicx} 
\usepackage{subfigure} 

\usepackage{natbib}

\usepackage{algorithm}
\usepackage{algorithmic}

\usepackage{multirow}
\usepackage{booktabs}

\usepackage{hyperref}
\usepackage{footmisc}

\usepackage[accepted]{arxiv}

\icmltitlerunning{DeCAF: A Deep Convolutional Activation Feature for Generic Visual Recognition}

\begin{document} 

\twocolumn[
\icmltitle{DeCAF: A Deep Convolutional Activation Feature \\ for Generic Visual Recognition}

\icmlauthor{Jeff Donahue$^*$, Yangqing Jia$^*$, Oriol Vinyals, Judy Hoffman, Ning Zhang, Eric Tzeng, Trevor Darrell}{\{jdonahue,jiayq,vinyals,jhoffman,nzhang,etzeng,trevor\}@eecs.berkeley.edu}
\icmladdress{UC Berkeley \& ICSI,
            Berkeley, CA, USA}

\icmlkeywords{machine learning, ICML}

\vskip 0.3in
]


\begin{abstract}

We evaluate whether features extracted from the activation of a deep
convolutional network trained in a fully supervised fashion on a
large, fixed set of object recognition tasks can be re-purposed to
novel generic tasks.  Our generic tasks may differ significantly from
the originally trained tasks and there may be insufficient labeled or
unlabeled data to conventionally train or adapt a deep architecture to
the new tasks.  We investigate and visualize the semantic clustering
of deep convolutional features with respect to a variety of such
tasks, including scene recognition, domain adaptation, and
fine-grained recognition challenges.  
We compare the efficacy of
relying on various network levels to define a fixed feature, and
report novel results that significantly outperform the
state-of-the-art on several important vision challenges.  We are releasing
DeCAF, an open-source implementation of these deep convolutional
activation features, along with all associated
network parameters to enable vision researchers to be able to conduct 
experimentation with deep representations across a range of visual
concept learning paradigms.

\end{abstract}

\section{Introduction}

Discovery of effective representations that capture salient semantics
for a given task is a key goal of perceptual learning.  Performance
with conventional visual representations, based on flat feature
representations involving quantized gradient filters, has been
impressive but has likely plateaued in recent years.

It has long been argued that deep or layered compositional
architectures should be able to capture salient aspects of a given
domain through discovery of salient clusters, parts, mid-level
features, and/or hidden units
\cite{autoencoders,fidler,yuille2007,efros2012,supervision}. Such models have
been able to perform better than traditional hand-engineered
representations in many domains, especially those where good features
have not already been engineered \cite{ngunsupervised}.  Recent
results have shown that moderately deep unsupervised models outperform
the state-of-the art gradient histogram features in part-based
detection models \cite{hsc}.

Deep models have recently been applied to large-scale
visual recognition tasks, trained via back-propagation through layers of convolutional filters
\citep{lecun89}. These models perform extremely well in domains with
large amounts of training data, and had early success in digit
classification tasks \cite{lenet}. With the advent of large scale
sources of category-level training data, e.g., \cite{imagenet_cvpr09}, and
efficient implementation with on-line approximate model averaging
(``dropout'') \cite{supervision}, they have recently outperformed all
known methods on a large scale recognition challenge
\cite{ilsvrc2012}.

With limited training data, however, fully-supervised deep
architectures with the representational capacity of \cite{supervision}
will generally dramatically overfit the training data.
In fact, many conventional visual recognition challenges have tasks with few
training examples; e.g., when a user is defining a category
``on-the-fly'' using specific examples, or for fine-grained
recognition challenges \cite{birds}, attributes \cite{attribute}, and/or
domain adaptation \cite{eccv_saenko}.


In this paper we investigate semi-supervised multi-task learning of
deep convolutional representations, where representations are learned
on a set of related problems but applied to new tasks which have too
few training examples to learn a full deep representation.  Our model
can either be considered as a deep architecture for transfer learning
based on a supervised pre-training phase, or simply as a new visual
feature \textit{DeCAF} defined by the convolutional network weights learned on a set
of pre-defined object recognition tasks.  Our work is also related to representation
learning schemes in computer vision which form an intermediate representation based
on learning classifiers on related tasks \cite{li2010object,torresani2010efficient,quattonidarrell}.

Our main result is the empirical validation that a generic visual
feature based on a convolutional network weights trained on ImageNet
outperforms a host of conventional visual representations on standard
benchmark object recognition tasks, including Caltech-101~\citep{caltech101},
the Office domain adaptation dataset~\citep{eccv_saenko},
the Caltech-UCSD Birds fine-grained recognition dataset~\citep{birds},
and the SUN-397 scene recognition database~\citep{xiao10}.

Further, we analyze the semantic salience of deep convolutional
representations, comparing visual features defined from such networks
to conventional representations.  In Section \ref{sec:decaf}, we visualize the
semantic clustering properties of deep convolutional features compared
to baseline representations, and find that convolutional features
appear to cluster semantic topics more readily than conventional
features.
Finally, while conventional deep learning can be computationally
expensive, we note that the run-time and resource computation of
deep-learned convolutional features are not exceptional in
comparison to existing features such as HOG~\cite{hog} or KDES~\citep{kdes}.

\section{Related work}

Deep convolutional networks have a long history in computer vision, with early examples showing successful results on using supervised back-propagation networks to perform digit recognition~\citep{lecun89}.
More recently, these networks, in particular the convolutional network proposed by~\citet{supervision}, have achieved competition-winning numbers on large benchmark datasets consisting of more than one million images, such as ImageNet~\citep{ilsvrc2012}.


Learning from related tasks also has a long history in machine learning beginning with~\citet{caruana1997multitask} and~\citet{thrun1996learning}. Later works such as~\citet{argyriou2008convex} developed efficient frameworks for optimizing representations from related tasks, and~\citet{ando2005framework} explored how to transfer parameter manifolds to new tasks.
In computer vision, forming a representation based on sets of trained classifiers on related tasks has recently been shown to be effective in a variety of retrieval and classification settings, specifically using classifiers based on visual category detectors~\citep{torresani2010efficient,li2010object}. A key question for such learning problems is to find a feature representation that captures the object category related information while discarding noise irrelevant to object category information such as illumination.

Transfer  learning across tasks using deep representations has been extensively studied, especially in an unsupervised setting~\citep{raina2007self,mesnil2012unsupervised}. However, reported successes with such models in convolutional networks have been limited to relatively small datasets such as CIFAR and MNIST, and efforts on larger datasets have had only modest success~\cite{googlebrain}.  We investigate the ``supervised pre-training'' approach proven successful in computer vision and multimedia settings using a concept-bank paradigm \citep{lscom,li2010object,torresani2010efficient} by learning the features on large-scale data in a supervised setting, then transferring them to different tasks with different labels.

To evaluate the generality of a representation formed from a deep convolutional feature trained on generic recognition tasks, we consider training and testing on datasets known to have a degree of dataset bias with respect to ImageNet. We evaluate on the SUN-397 scene dataset, as well as datasets used to evaluate domain adaptation performance directly~\citep{ref:dlid,ref:kulis_cvpr11}. This evaluates whether the learned features could undo the domain bias by capturing the real semantic information instead of overfitting to domain-specific appearances.



\section{Deep Convolutional Activation Features}
\label{sec:decaf}

\newcommand{\fsize}{.24\linewidth}
\begin{figure*}[t]
\centering
	\subfigure[LLC]{\includegraphics[width=\fsize]{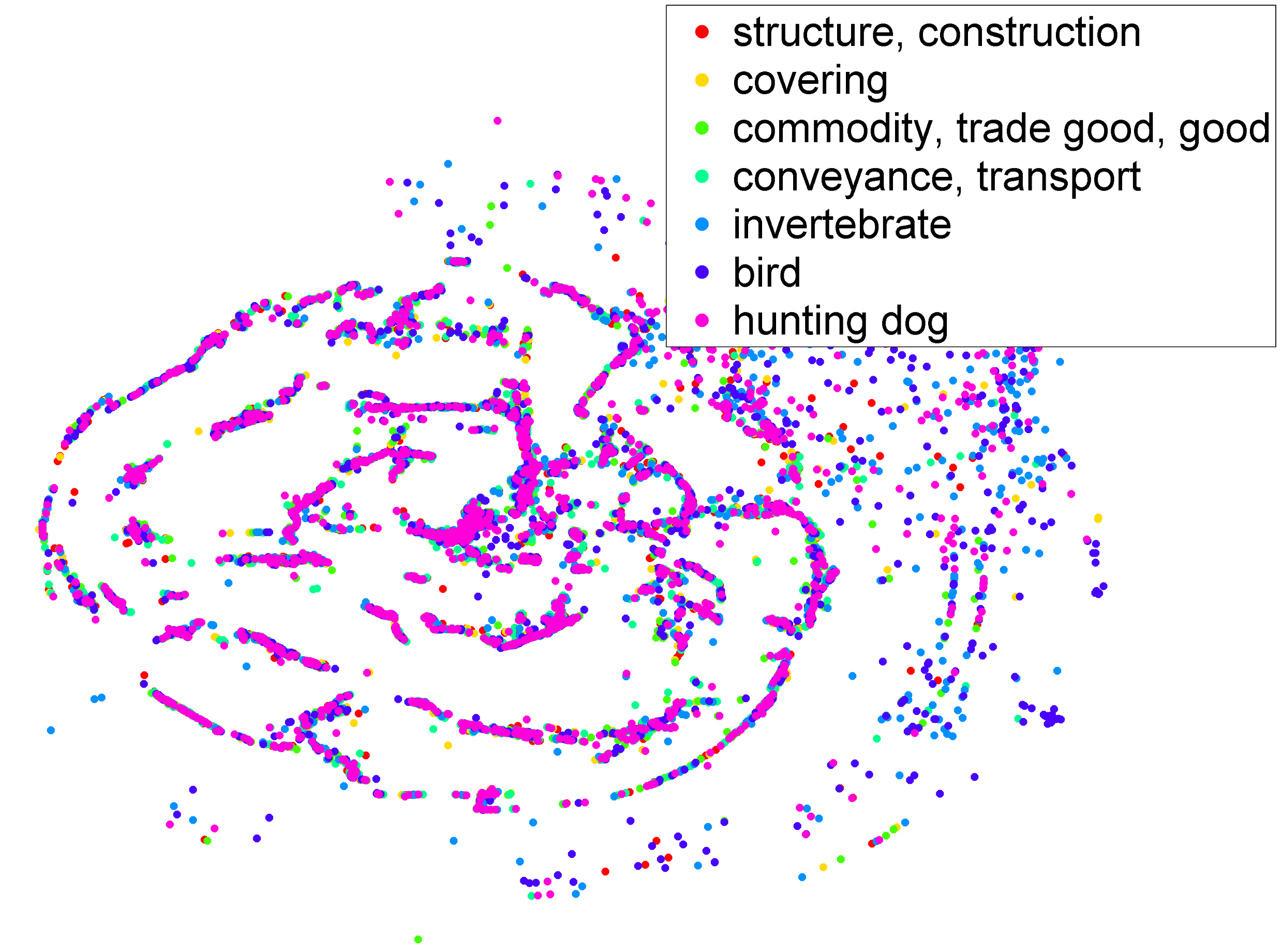}}
	\subfigure[GIST]{\includegraphics[width=\fsize]{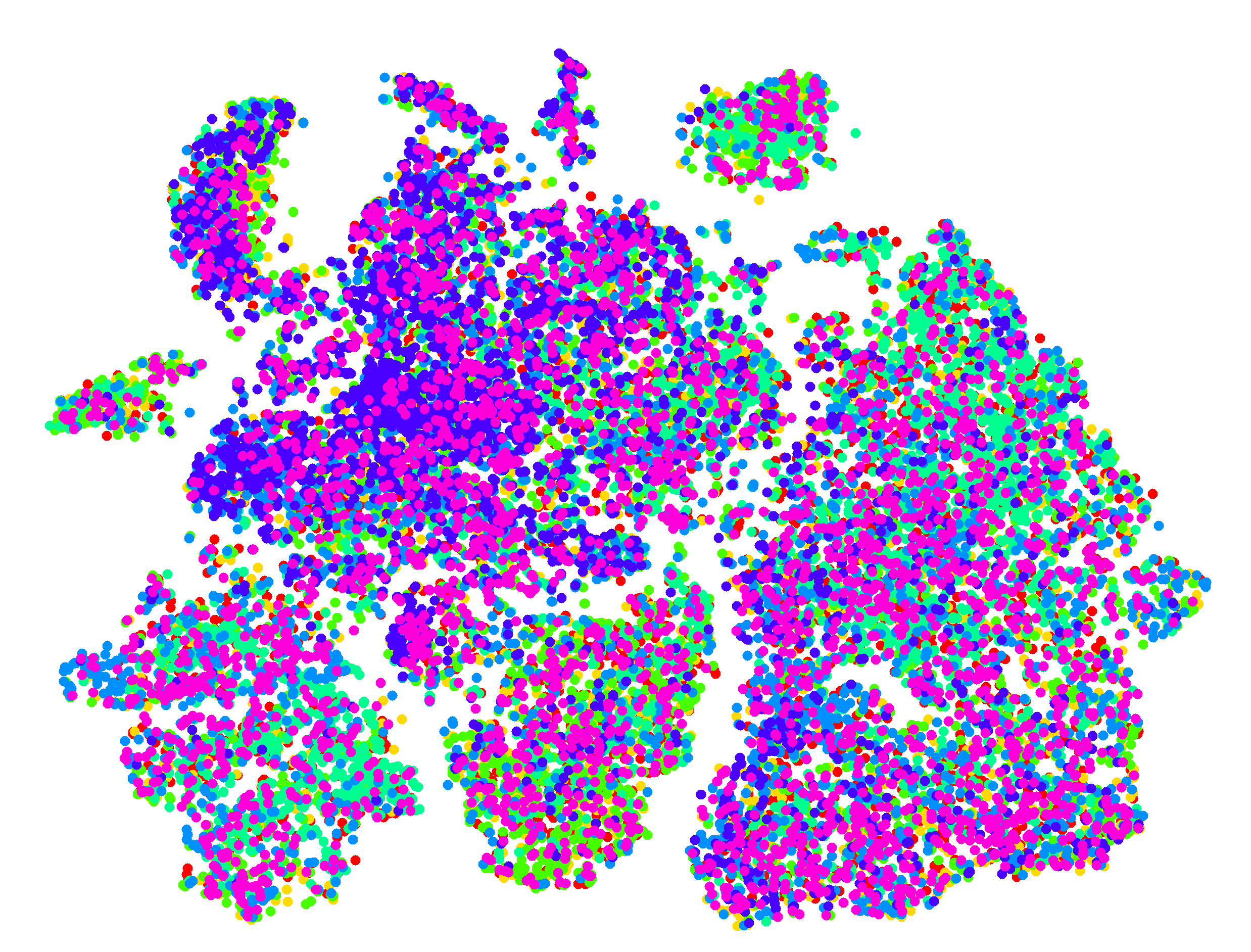}}
	\subfigure[DeCAF$_1$]{\includegraphics[width=\fsize]{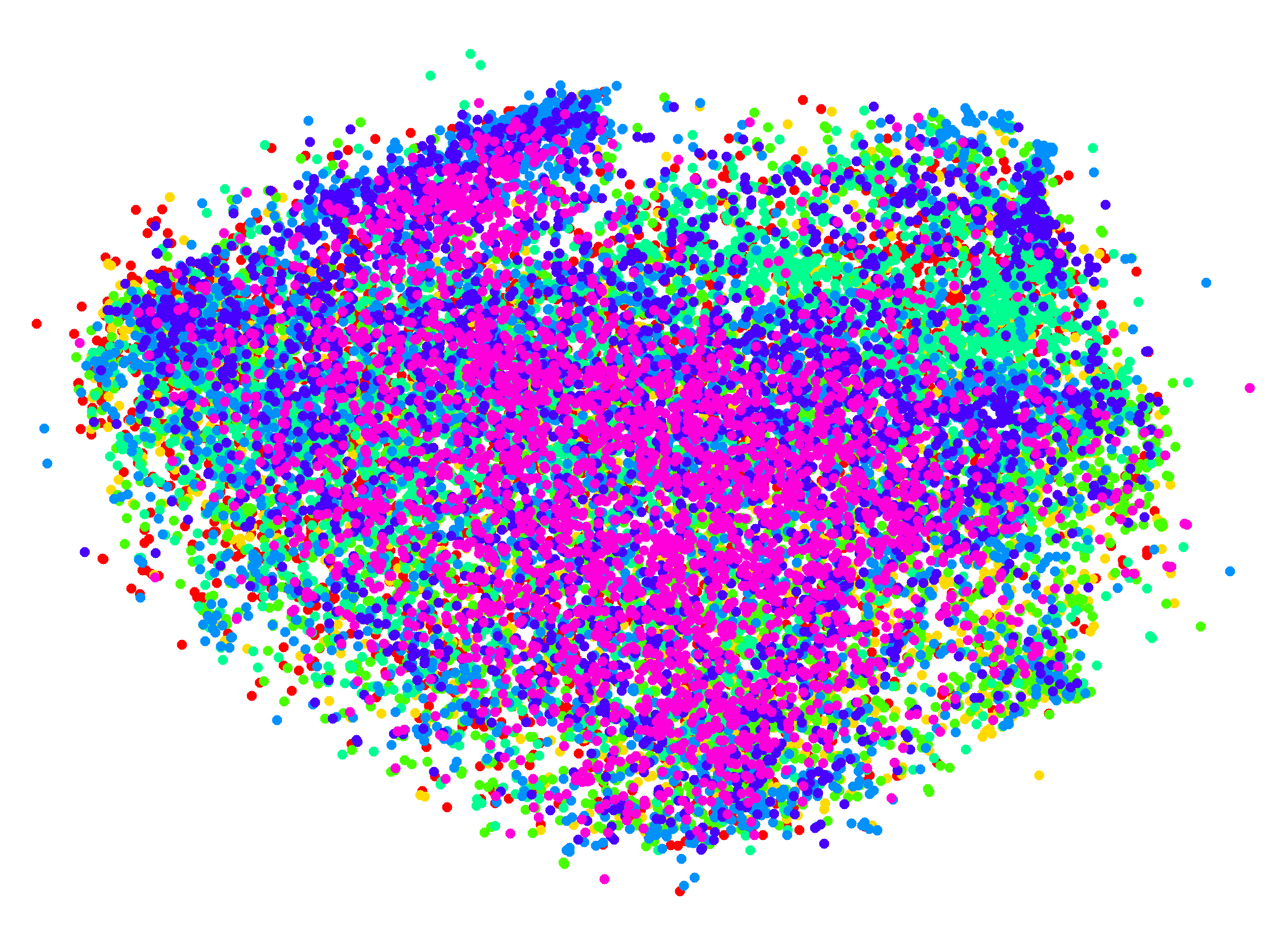}}
	\subfigure[DeCAF$_6$]{\includegraphics[width=\fsize]{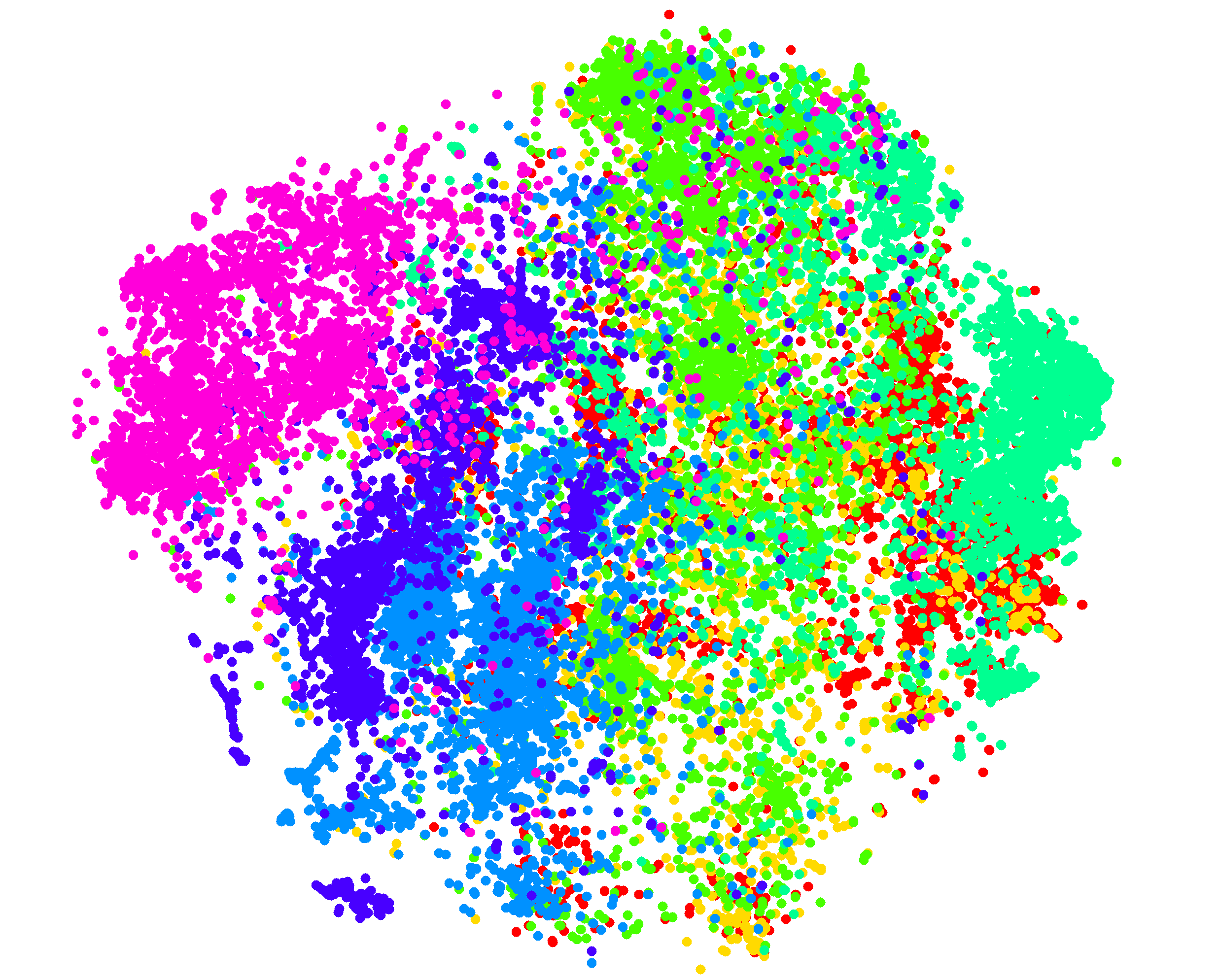}}
\caption{This figure shows several t-SNE feature visualizations on the ILSVRC-2012 validation set. (a) LLC , (b) GIST, and features derived from our CNN: (c) DeCAF$_1$, the first pooling layer, and (d) DeCAF$_6$, the second to last hidden layer (best viewed in color). \label{fig:comparison}}
\end{figure*}

In our approach, a deep convolutional model is first trained in a fully supervised setting using a state-of-the-art method \citet{supervision}.  We then extract various features from this network, and evaluate the efficacy of these features on generic vision tasks.   
Even though the forward pass computed by the architecture in this section does achieve state-of-the-art performance on ILSVRC-2012, two questions remain:
\begin{itemize}\setlength{\itemsep}{0pt}
\item Do features extracted from the CNN generalize to other datasets?
\item How do these features perform versus depth?
\end{itemize}
We address these questions both qualitatively and quantitatively, via visualizations of semantic clusters below, and experimental comparision to current baselines in the following section. 

\subsection{An Open-source Convolutional Model}
To facilitate the wide-spread analysis of deep convolutional features, we developed a Python framework that allows one to easily train networks consisting of various layer types and to execute pre-trained networks efficiently without being restricted to a GPU (which in many cases may hinder the deployment of trained models). Specifically, we adopted open-source Python packages such as {\tt numpy/scipy} for efficient numerical computation, with parts of the computation-heavy code implemented in C and linked to Python. In terms of computation speed, our model is able to process about 40 images per second with an 8-core commodity machine when the CNN model is executed in a minibatch mode.

Our implementation, \texttt{decaf}, will be publicly available\footnote{\url{http://decaf.berkeleyvision.org/}}. In addition, we will release the network parameters used in our experiments to allow for out-of-the-box feature extraction without the need to re-train the large network\footnote{We note that although our CPU implementation allows one to also train networks, that training of large networks such as the ones for ImageNet may still be time-consuming on CPUs, and we rely on our own implementation of the network by extending the \texttt{cuda-convnet} GPU framework provided by Alex Krizhevsky to train such models.}.  This also aligns with the philosophy of supervised transfer: one may view the trained model as an analog to the prior knowledge a human obtains from previous visual experiences, which helps in learning new tasks more efficiently.

As the underlying architecture for our feature we adopt the deep convolutional neural network architecture proposed by~\citet{supervision}, which won the ImageNet Large Scale Visual Recognition Challenge 2012~\citep{ilsvrc2012} with a top-1 validation error rate of 40.7\%.
\footnote{The model entered into the competition actually achieved a top-1 validation error rate of 36.7\% by averaging the predictions of 7 structurally identical models that were initialized and trained independently.  We trained only a single instance of the model; hence we refer to the single model error rate of 40.7\%.}
We chose this model due to its performance on a difficult 1000-way classification task, hypothesizing that the activations of the neurons in its late hidden layers might serve as very strong features for a variety of object recognition tasks.
Its inputs are the mean-centered raw RGB pixel intensity values of a $224\times224$ image. These values are forward propagated through 5 convolutional layers (with pooling and ReLU non-linearities applied along the way) and 3 fully-connected layers to determine its final neuron activities: a distribution over the task's 1000 object categories.
Our instance of the model attains an error rate of \textbf{42.9\%} on the ILSVRC-2012 validation set -- 2.2\% shy of the 40.7\% achieved by~\cite{supervision}.



We refer to~\citet{supervision} for a detailed discussion of the architecture and training protocol, which we closely followed with the exception of two small differences in the input data.
First, we ignore the image's original aspect ratio and warp it to $256\times256$, rather than resizing and cropping to preserve the proportions.
Secondly, we did not perform the data augmentation trick of adding random multiples of the principle components of the RGB pixel values throughout the dataset, proposed as a way of capturing invariance to changes in illumination 
 and color\footnote{According to the authors, this scheme reduced their models' test set error by over 1\%, likely explaining much of our network's performance discrepancy.}.

\subsection{Feature Generalization and Visualization}

We visualized the model features to gain insight into the semantic capacity of DeCAF and other features that have been typically employed in computer vision.
In particular, we compare the features described in Section~\ref{sec:decaf} with GIST features \cite{gist} and LLC features \cite{llc}.

We visualize features in the following way: we run the t-SNE algorithm \cite{tsne} to find a 2-dimensional embedding of the high-dimensional feature space, and plot them as points colored depending on their semantic category in a particular hierarchy. We did this on the validation set of ILSVRC-2012 to avoid overfitting effects (as the deep CNN used in this paper was trained only on the training set), and also use an independent dataset, SUN-397 \cite{xiao10}, to evaluate how dataset bias affects our results (see e.g. \cite{torralba_cvpr11} for a deeper discussion of this topic).

One would expect features closer to the output (softmax) layer to be
linearly separable, so it is not very interesting (and also
visually quite hard) to represent the 1000 classes on the t-SNE
derived embedding. 

We first visualize the semantic segregation of the model by plotting
the embedding of labels for higher levels of the WordNet hierarchy;
for example, a strong feature for visual recognition should cluster
indoor and outdoor instances separately, even though there is no
explicit modeling through the supervised training of the CNN.
 Figure~\ref{fig:comparison} shows the features extracted on the validation set using the first pooling layer, and the second to last fully connected layer, showing a clear semantic clustering in the latter but not in the former. This is compatible with common deep learning knowledge that the first layers learn ``low-level'' features, whereas the latter layers learn semantic or ``high-level'' features. Furthermore, other features such as GIST or LLC fail to capture the semantic difference in the image (although they show interesting clustering structure).\footnote{Some of the features were very high dimensional (e.g. LLC had 16K dimension), in which case we preprocess them by randomly projecting them down to 512 dimensions -- random projections are cheap to apply and tend to preserve distances well, which is all the t-SNE algorithm cares about.}

\begin{figure}
\centering
\includegraphics[width=.8\linewidth]{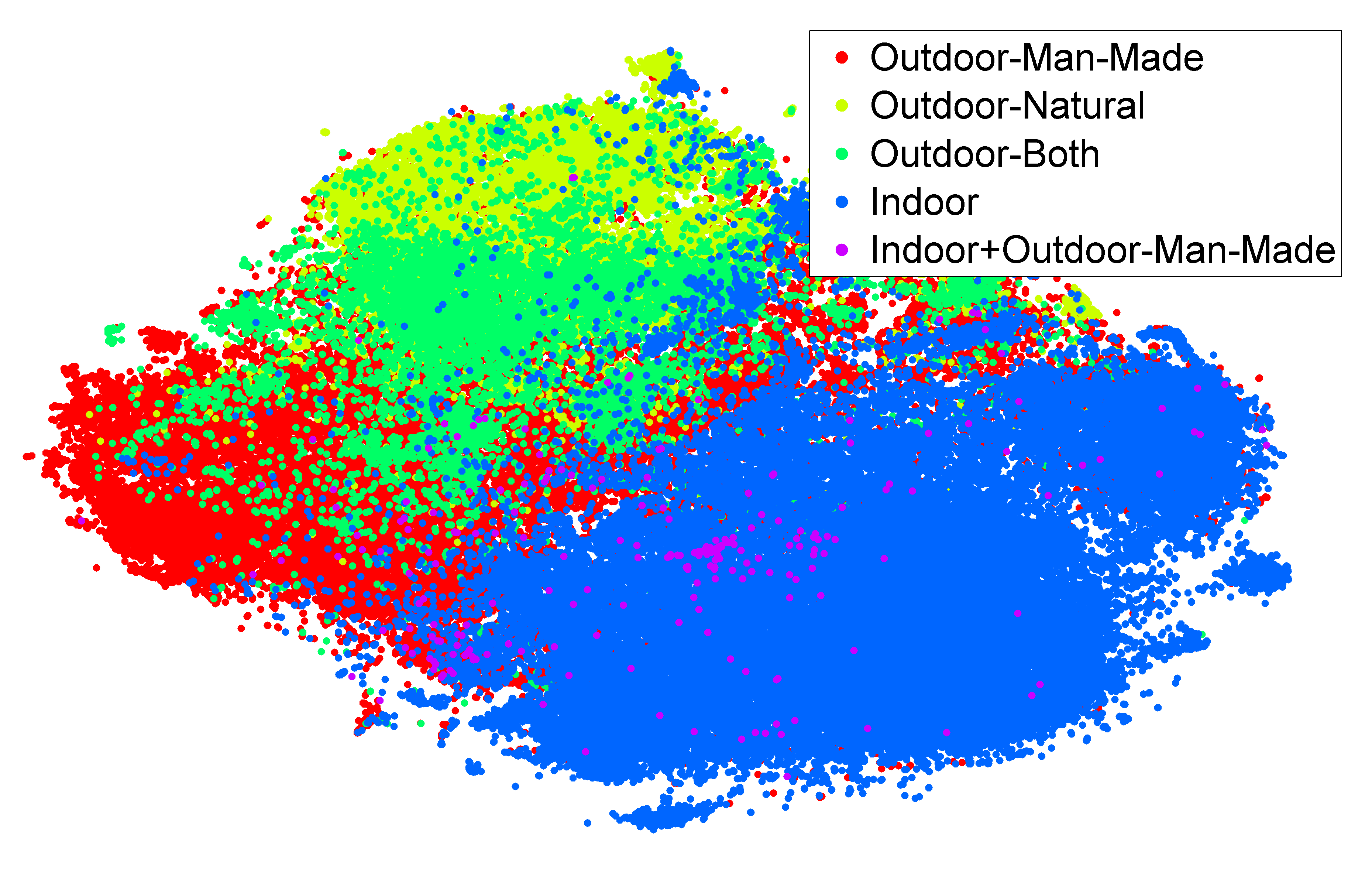}
\caption{In this figure we show how our features trained on ILSVRC-2012 generalized to SUN-397 when considering semantic groupings of labels (best viewed in color). \label{fig:generalization}}
\vspace{-.5cm}
\end{figure}

More interestingly, in Figure~\ref{fig:generalization} we can see the top performing features (DeCAF$_6$) on the SUN-397 dataset. Even there, the features show very good clustering of semantic classes (e.g., indoor vs. outdoor). This suggests DeCAF is a good feature for general object recognition tasks. Consider the case where the object class that we are trying to detect is not in the original object pool of ILSVRC-2012. The fact that these features cluster several intermediate nodes of WordNet implies that these features are an excellent starting point for generalizing to unseen classes.

\subsection{Time Analysis}

While it is generally believed that convolutional neural networks take a significant
amount of time to execute, a detailed analysis of the computation time over the multiple
layers involved is still missing in the literature. In this subsection we report a
break-down of the computation time analyzed using the \texttt{decaf} framework.

\begin{figure}
\centering
\subfigure[]{
\includegraphics[width=0.6\linewidth]{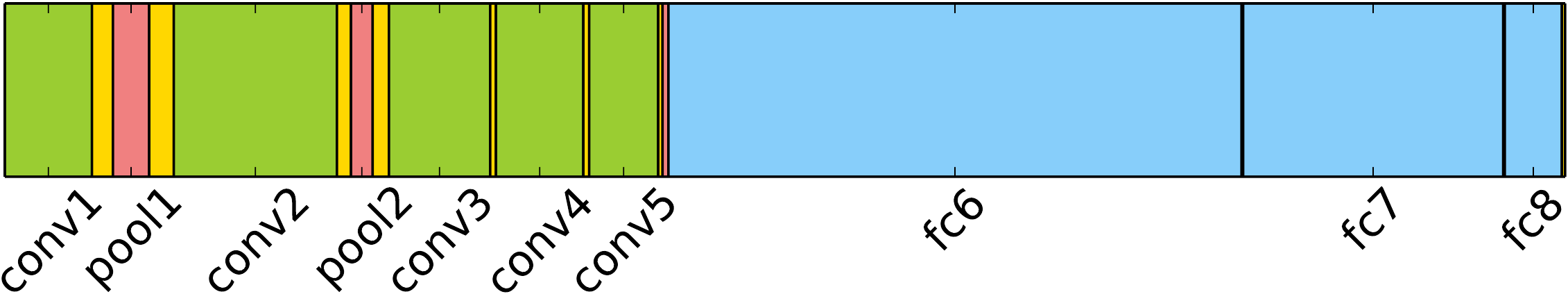}
\label{fig:time:sequence}
}
\subfigure[]{
\includegraphics[width=0.34\linewidth]{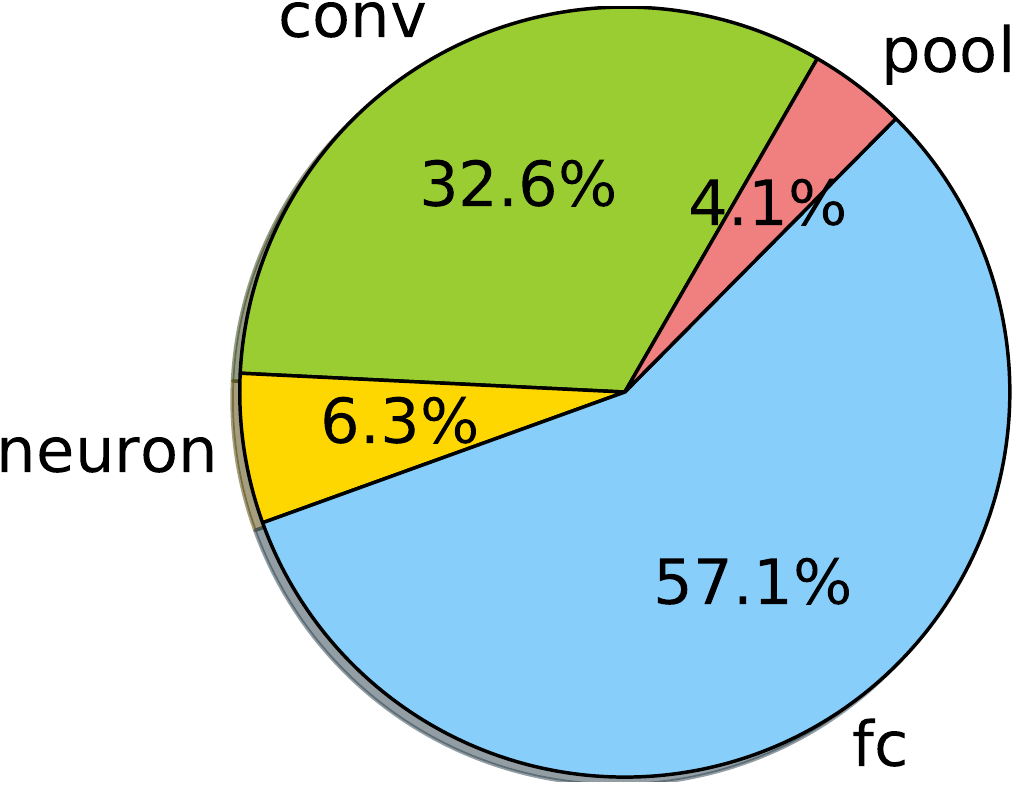}
\label{fig:time:piechart}}
\caption{(a) The computation time on each layer when running classification on one single input image. The layers with the most time consumption are labeled. (b) The distribution of computation time over different layer types. In the piechart, fc = fully connected layers, conv = convolution layers, pool = pooling layers, and neuron = neuron layers such as ReLU, sigmoid, and dropout.}
\end{figure}

In Figure \ref{fig:time:sequence} we lay out the computation time spent on individual
layers with the most time-consuming layers labeled. We observe that the convolution and fully-connected layers take most of the time to run, which is understandable as they involve large matrix-matrix multiplications\footnote{We implemented the convolutional layers as an {\tt im2col} step followed by dense matrix multiplication, which empirically worked best with small kernel sizes and large number of kernels.}. Also, the time distribution over different layer types (Figure \ref{fig:time:piechart}) reveals an
interesting fact: in large networks such as the current ImageNet CNN model, the last few fully-connected layers require the most computation time as they involve large transform matrices. This is particularly important when one considers classification into a larger number of categories or with larger hidden-layer sizes, suggesting that certain sparse approaches such as Bayesian output coding \cite{hsu2009multi} may be necessary to carry out classification into even larger number of object categories.

\section{Experiments}
\label{sec:experiments}

\begin{figure*}[t]
\centering
\label{fig:caltech101results}
\begin{minipage}{0.60\textwidth}
\small{
\begin{tabular}{lccc}

\toprule

& DeCAF$_5$ & DeCAF$_6$ & DeCAF$_7$ \\
\midrule
LogReg & $63.29 \pm 6.6$ & $84.30 \pm 1.6$ & $84.87 \pm 0.6$ \\
LogReg with Dropout & - & $86.08 \pm 0.8$ & $85.68 \pm 0.6$ \\
SVM & $77.12 \pm 1.1$ & $84.77 \pm 1.2$ & $83.24 \pm 1.2$ \\
SVM with Dropout & - & $\mathbf{86.91 \pm 0.7}$ & $85.51 \pm 0.9$ \\
\\
\citet{yang09} & \multicolumn{3}{c}{84.3} \\
\citet{jarrett09} & \multicolumn{3}{c}{65.5} \\
\bottomrule 
\end{tabular}
}
\end{minipage}%
\begin{minipage}{0.40\textwidth}
\hfill \includegraphics[width=0.8\textwidth]{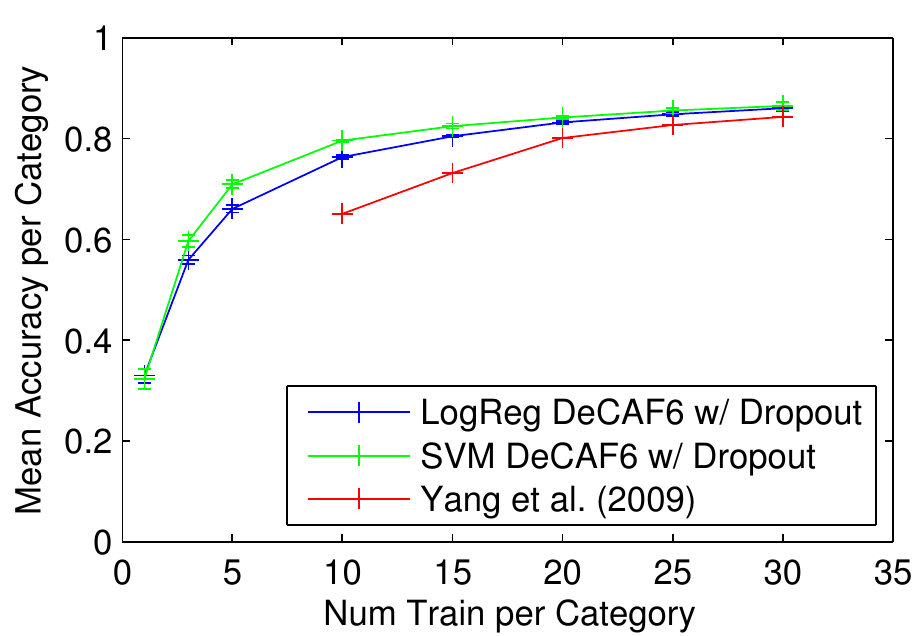}
\end{minipage}
\caption{Left: average accuracy per class on Caltech-101 with 30 training samples per class across three hidden layers of the network and two classifiers.
Our result from the training protocol/classifier combination with the best validation accuracy -- SVM with Layer 6 (+ dropout) features -- is shown in \textbf{bold}.
Right: average accuracy per class on Caltech-101 at varying training set sizes.
}
\end{figure*}

In this section, we present experimental results evaluating DeCAF on multiple standard computer vision benchmarks, comparing many possible featurization and classification approaches.
In each of the experiments, we take the activations of the $n^{\mathrm{th}}$ hidden layer of the deep convolutional neural network described in Section~\ref{sec:decaf} as a feature DeCAF$_n$.
DeCAF$_7$ denotes features taken from the final hidden layer -- i.e., just before propagating through the final fully connected layer to produce the class predictions.
DeCAF$_6$ is the activations of the layer before DeCAF$_7$, and DeCAF$_5$ the layer before DeCAF$_6$.
DeCAF$_5$ is the first set of activations that has been fully propagated through the convolutional layers of the network.
We chose not to evaluate features from any earlier in the network, as the earlier convolutional layers are unlikely to contain a richer semantic representation than the later features which form higher-level hypotheses from the low to mid-level local information in the activations of the convolutional layers.
Because we are investigating the use of the network's hidden layer activations as features, all of its weights are frozen to those learned on the~\citet{ilsvrc2012} dataset.\footnote{We also experimented with the equivalent feature using randomized weights and found it to have performance comparable to traditional hand-designed features.}
All images are preprocessed using the procedure described for the ILSVRC images in Section~\ref{sec:decaf}, taking features on the center $224 \times 224$ crop of the $256 \times 256$ resized image.

We present results on multiple datasets to evaluate the strength of DeCAF for basic object recognition, domain adaptation, fine-grained recognition, and scene recognition.
These tasks each differ somewhat from that for which the architecture was trained, together representing much of the contemporary visual recognition spectrum.

\subsection{Object recognition}
\label{sec:caltech}
To analyze the ability of the deep features to transfer to basic-level object category recognition, we evaluate them on the Caltech-101 dataset~\citep{caltech101}.
In addition to directly evaluating linear classifier performance on DeCAF$_6$ and DeCAF$_7$, we also report results using a regularization technique called ``dropout'' proposed by~\citet{hintondropout}.
At training time, this technique works by randomly setting half of the activations (here, our features) in a given layer to 0.
At test time, all activations are multiplied by 0.5.
Dropout was used successfully by~\citet{supervision} in layers 6 and 7 of their network; hence we study the effect of the technique when applied to the features derived from these layers.

In each evaluation, the classifier, a logistic regression (LogReg) or support vector machine (SVM), is trained on a random set of 30 samples per class (including the background class), and tested on the rest of the data, with parameters cross-validated for each split on a 25 train/5 validation subsplit of the training data.
The results in Figure~\ref{fig:caltech101results}, left, are reported in terms of mean accuracy per category averaged over five data splits.

Our top-performing method (based on validation accuracy) trains a linear SVM on DeCAF$_6$ with dropout, with test set accuracy of 86.9\%.
The DeCAF$_5$ features perform substantially worse than either the DeCAF$_6$ or DeCAF$_7$ features, and hence we do not evaluate them further in this paper.
The DeCAF$_7$ features generally have accuracy about 1-2\% lower than the DeCAF$_6$ features on this task.
The dropout regularization technique uniformly improved results by 0-2\% for each classifier/feature combination.
When trained on DeCAF, the SVM and logistic regression classifiers perform roughly equally well on this task.

We compare our performance against the current state-of-the-art on this benchmark from~\citet{yang09}, a method employing a combination of 5 traditional hand-engineered image features followed by a multi-kernel based classifier.
Our top-performing method training a linear SVM on a single feature outperforms this method by 2.6\%.
Our method also outperforms by over 20\% the two-layer convolutional network of~\citet{jarrett09}, demonstrating the importance of the depth of the network used for our feature.
Note that unlike our method, these approaches from the literature do not implicitly leverage an outside large-scale image database like ImageNet.
The performance edge of our method over these approaches demonstrates the importance of multi-task learning when performing object recognition with sparse data like that available in the Caltech-101 benchmark.

We also show how performance of the two DeCAF$_6$ with dropout methods above vary with the number of training cases per category, plotted in Figure~\ref{fig:caltech101results}, right, trained with fixed parameters and evaluated under the same metric as before.
Our one-shot learning results (e.g., 33.0\% for SVM) suggest that with sufficiently strong representations like DeCAF, useful models of visual categories can often be learned from just a single positive example.


\subsection{Domain adaptation}
We next evaluate DeCAF for use on the task of domain adaptation. For our experiments we use the benchmark \textit{Office} dataset \citep{eccv_saenko}.The dataset contains three domains: \texttt{Amazon}, which consists of product images taken from \url{amazon.com}; and \texttt{Webcam} and \texttt{Dslr}, which consists of images taken in an office environment using a webcam or digital SLR camera, respectively. 

In the domain adaptation setting, we are given a training (source) domain with labeled training data and a distinct test (target) domain with either a small amount of labeled data or no labeled data. We will experiment within the supervised domain adaptation setting, where there is a small amount of labeled data available from the target domain. 

Most prior work for this dataset uses SURF \citep{surf06} interest point features (available for download with the dataset). To illustrate the ability of DeCAF to be robust to resolution changes, we  use the t-SNE~\citep{tsne} algorithm to project both SURF and DeCAF$_6$, computed for \texttt{Webcam} and \texttt{Dslr}, into a 2D visualizable space (See Figure~\ref{fig:office_vis}). We visualize an image on the point in space corresponding to its low dimension projected feature vector. We find that DeCAF not only provides better within category clustering, but also clusters same category instances across domains. This indicates qualitatively that DeCAF removed some of the domain bias between the \texttt{Webcam} and \texttt{Dslr} domains.
\newcommand{\soffice}{.7\linewidth}
\begin{figure}[t!]
\centering
\subfigure[SURF features]{
\includegraphics[width=\soffice, height=\soffice]{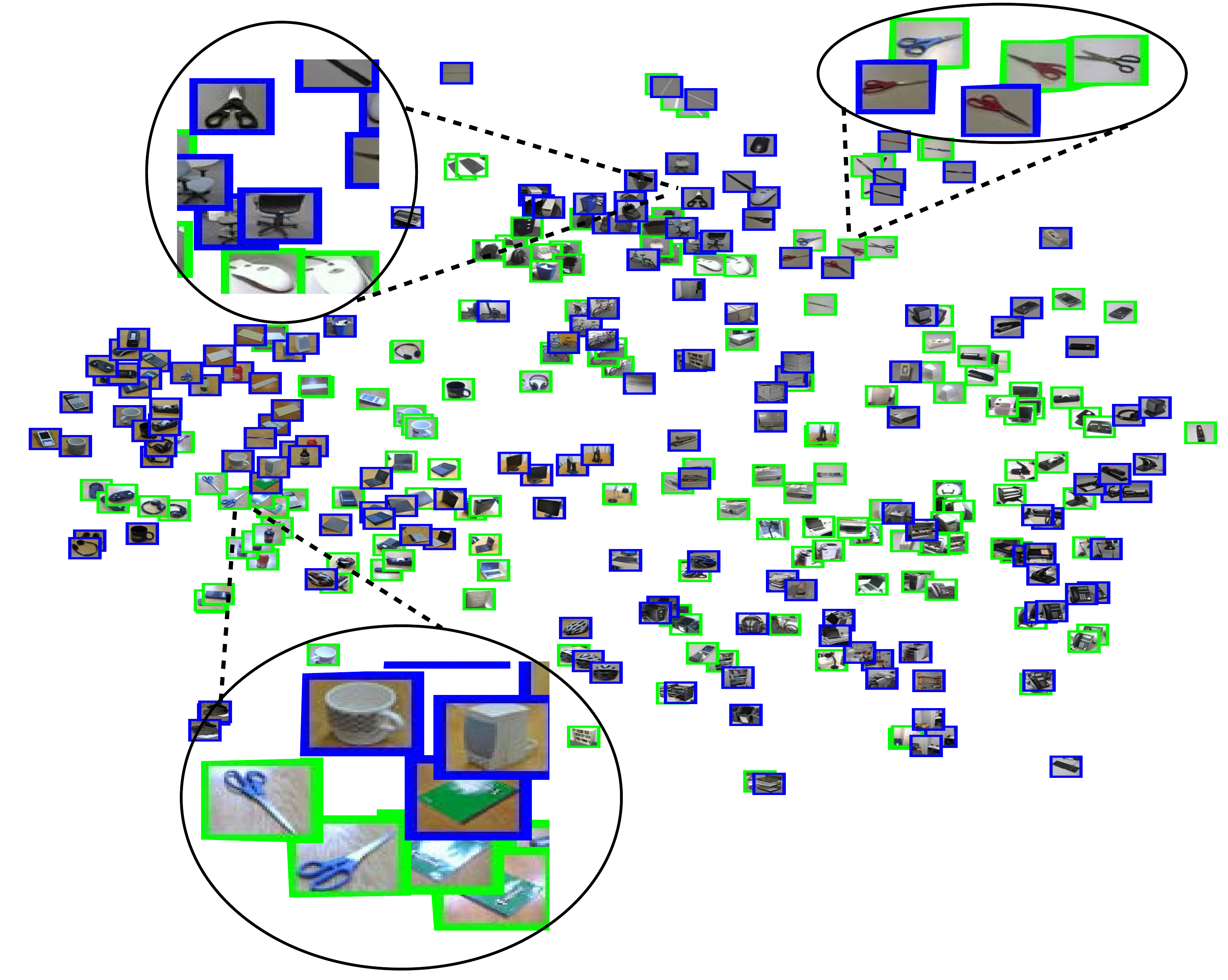}}\\ 
\subfigure[DeCAF$_6$]{
\includegraphics[width=\soffice, height=\soffice]{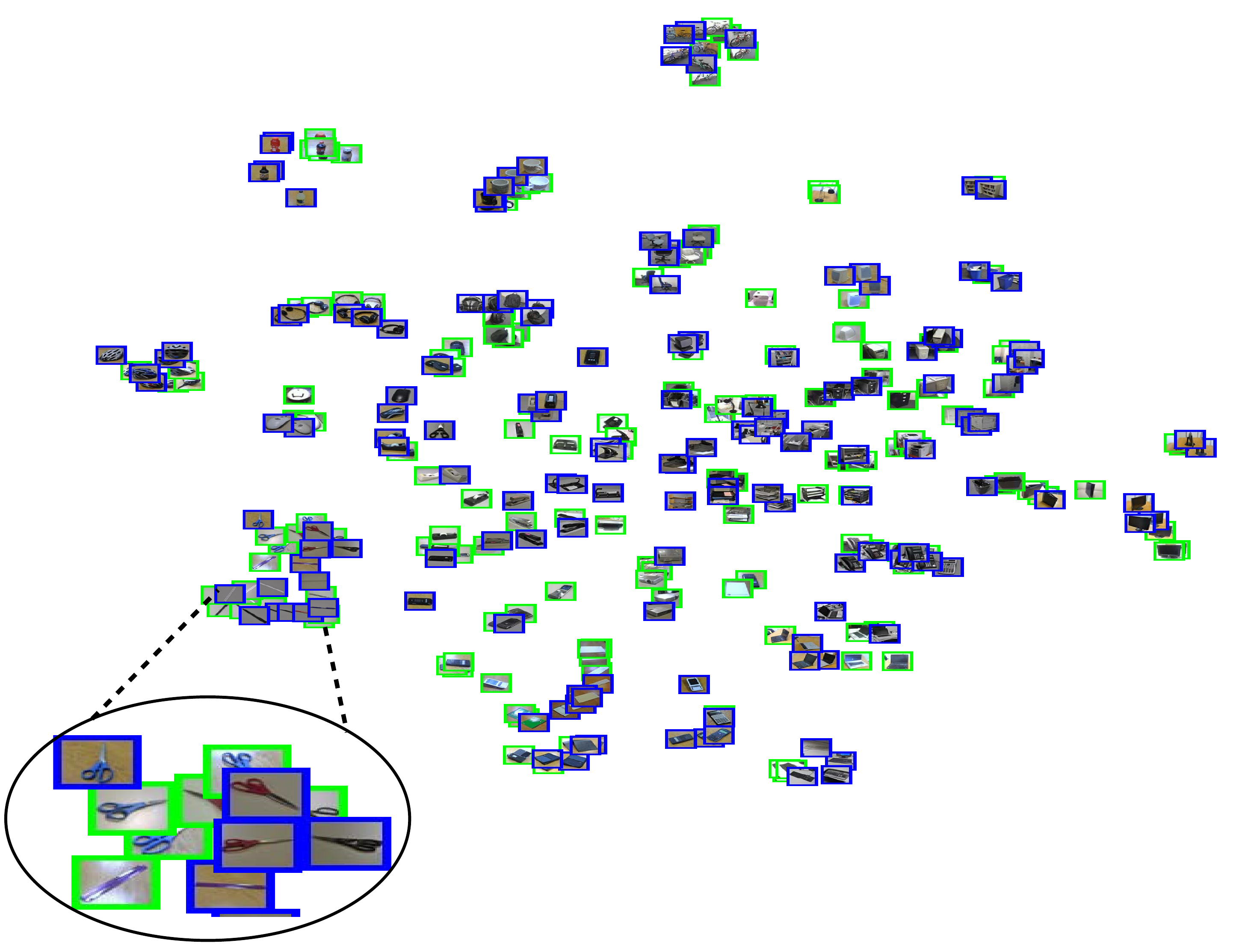}
}
\caption{Visualization of the webcam (green) and dslr (blue) domains using the original released SURF features (a) and DeCAF$_6$ (b). The figure is best viewed by zooming in to see the images in local regions. All images from the scissor class are shown enlarged. They are well clustered and overlapping in both domains with our representation, while SURF only clusters a subset and places the others in disjoint parts of the space, closest to distinctly different categories such as chairs and mugs.}
\label{fig:office_vis}
\end{figure}

We validate this conclusion with a quantitative experiment on the \textit{Office} dataset. Table \ref{tab:office} presents multi-class accuracy averaged across 5 train/test splits for the domain shifts \texttt{Amazon}$\rightarrow$\texttt{Webcam} and \texttt{Dslr} $\rightarrow$ \texttt{Webcam}. We use the standard experimental setup first presented in \citet{eccv_saenko}. To compare SURF with the DeCAF$_6$, and DeCAF$_7$ deep convolutional features, we report the multi-class accuracy for each, using an SVM and Logistic Regression both trained in 3 ways: with only source data (S), only target data (T), and source and target data (ST). We also report results for three adaptive methods run with each DeCAF we consider as input. Finally, for completeness we report a recent and competing deep domain adaptation result from \citet{ref:dlid}. DeCAF dramatically outperforms the baseline SURF feature available with the \textit{Office} dataset as well as the deep adaptive method of \citet{ref:dlid}. 

\newcommand{\ra}[1]{\renewcommand{\arraystretch}{#1}}

\begin{table*}
\small
\begin{center}
\ra{1.1}
\tabcolsep=0.11cm
\begin{tabular}{@{} lccccccc @{}}
\toprule
& \multicolumn{3}{c}{\texttt{Amazon} $\rightarrow$ \texttt{Webcam}} & \phantom{ab} & \multicolumn{3}{c}{\texttt{Dslr} $\rightarrow$ \texttt{Webcam}}\\
\cmidrule{2-4} \cmidrule{6-8}
	& SURF & DeCAF$_6$ & DeCAF$_7$  && SURF& DeCAF$_6$ & DeCAF$_7$\\
        	\midrule
Logistic Reg. (S) & $  9.63\pm    1.4$ & $ 48.58\pm    1.3$ & $ 53.56\pm    1.5$ && $ 24.22\pm    1.8 $ & $ 88.77\pm    1.2$ & $ 87.38\pm    2.2$ \\
SVM (S) & $ 11.05\pm    2.3$ & $ 52.22\pm    1.7$ & $ 53.90\pm    2.2$ && $ 38.80\pm    0.7 $ & $ 91.48\pm    1.5$ & $ 89.15\pm    1.7$ \\
\\
Logistic Reg. (T) & $ 24.33\pm    2.1$ & $ 72.56\pm    2.1$ & $ 74.19\pm    2.8$ && $ 24.33\pm    2.1 $& $ 72.56\pm    2.1$ & $ 74.19\pm    2.8$ \\
SVM (T) & $ 51.05\pm    2.0$ &  $ 78.26\pm    2.6$ & $ 78.72\pm    2.3$ && $ 51.05\pm    2.0 $&  $ 78.26\pm    2.6$ & $ 78.72\pm    2.3$ \\
\\
Logistic Reg. (ST) & $ 19.89\pm    1.7$  & $ 75.30\pm    2.0$ & $ 76.32\pm    2.0$ && $ 36.55\pm    2.2 $ & $ 92.88\pm    0.6$ & $ 91.91\pm    2.0$ \\
SVM (ST) & $ 23.19\pm    3.5$  & $ 80.66\pm    2.3$ & $ 79.12\pm    2.1$ && $ 46.32\pm    1.1 $&$\bm{ 94.79\pm    1.2}$ & $ 92.96\pm    2.0$ \\
 \\
\citet{ref:daume} & $ 40.26\pm    1.1$  & $\bm{82.14\pm    1.9}$ & $ 81.65\pm    2.4$ && $ 55.07\pm    3.0 $ & $ 91.25\pm    1.1$ & $ 89.52\pm    2.2$ \\
\citet{Hoffman13:ELD} & $ 37.66\pm    2.2$  & $ 80.06\pm    2.7$ & $ 80.37\pm    2.0$ && $ 53.65\pm    3.3 $ & $ 93.25\pm    1.5$ & $ 91.45\pm    1.5$ \\
\citet{ref:gong12_gfk} & $ 39.80\pm    2.3$ &  $ 75.21\pm    1.2$ & $ 77.55\pm    1.9$ & &$ 39.12\pm    1.3 $ & $ 88.40\pm    1.0$ & $ 88.66\pm    1.9$ \\
\\
\citet{ref:dlid} & \multicolumn{3}{c}{58.85} && \multicolumn{3}{c} {78.21}\\
\bottomrule 
\end{tabular} 
\end{center}
\caption{DeCAF dramatically outperforms the baseline SURF feature available with the \textit{Office} dataset as well as the deep adaptive method of \citet{ref:dlid}. We report average multi class accuracy using both non-adaptive and adaptive classifiers, changing only the input feature from SURF to DeCAF. Most surprisingly, in the case of \texttt{Dslr}$\rightarrow$\texttt{Webcam} the domain shift is largely non-existent with DeCAF.}
\label{tab:office}
\end{table*}

\subsection{Subcategory recognition}
\label{sec:birds}

We tested the performance of DeCAF on the task of subcategory recognition. To this end, we adopted one of its most popular tasks - the Caltech-UCSD birds dataset~\citep{birds}, and compare the performance against several state-of-the-art baselines.

Following common practice in the literature, we adopted two approaches to perform classification. Our first approach adopts an ImageNet-like pipeline, in which we followed the existing protocol by cropping the images regions $1.5\times$ the size of the provided bounding boxes, resizing them 256$\times$256 and then feeding them into the CNN pipeline to get the features for classification. We computed DeCAF$_6$ and trained a multi-class logistic regression on top of the features.

\begin{figure}[t]
\centering
\subfigure[DPM detections]{
\includegraphics[height=0.3\linewidth]{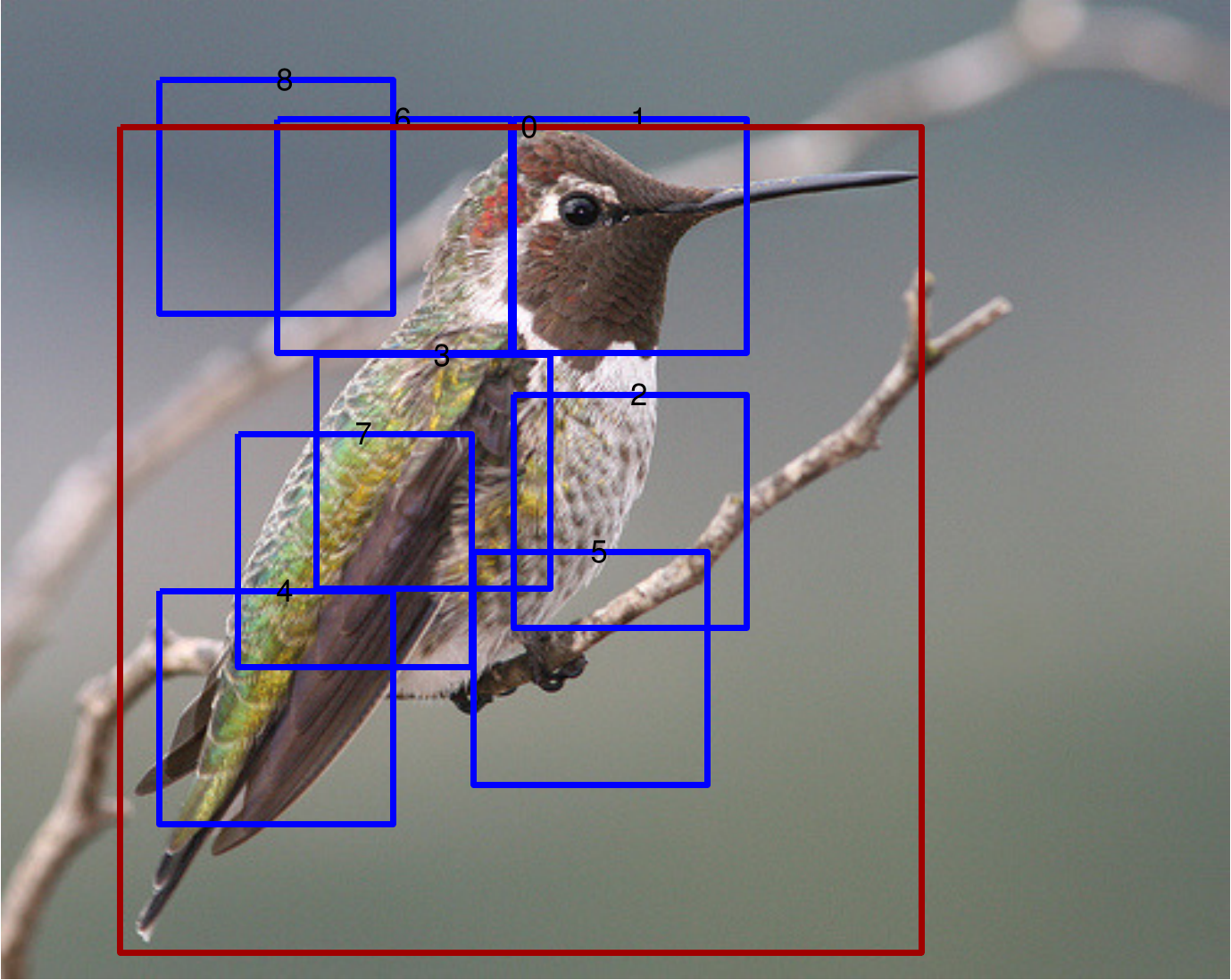}}\hfill
\subfigure[Parts]{ 
\includegraphics[height=0.3\linewidth]{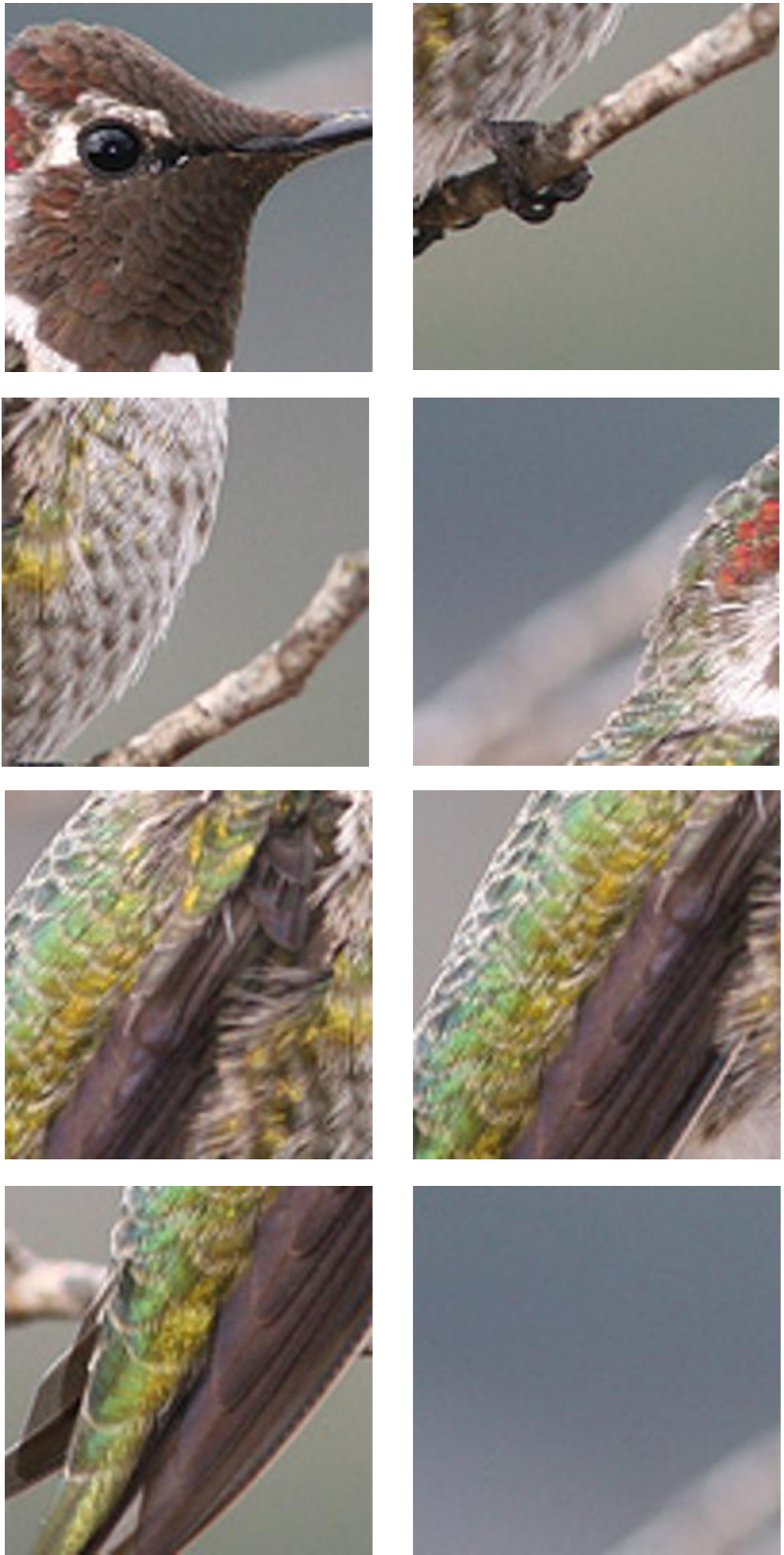}}\hfill
\subfigure[DPD]{%
\raisebox{10mm}{\begin{minipage}{.2\linewidth}
\centering
\includegraphics[height=0.47\linewidth]{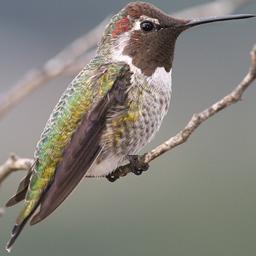} \\
\includegraphics[height=0.47\linewidth]{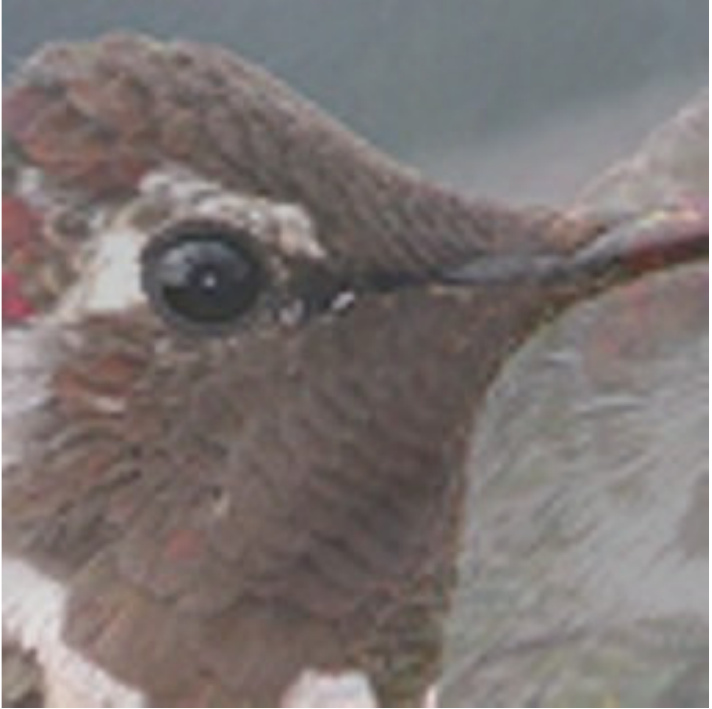} \\
\includegraphics[height=0.47\linewidth]{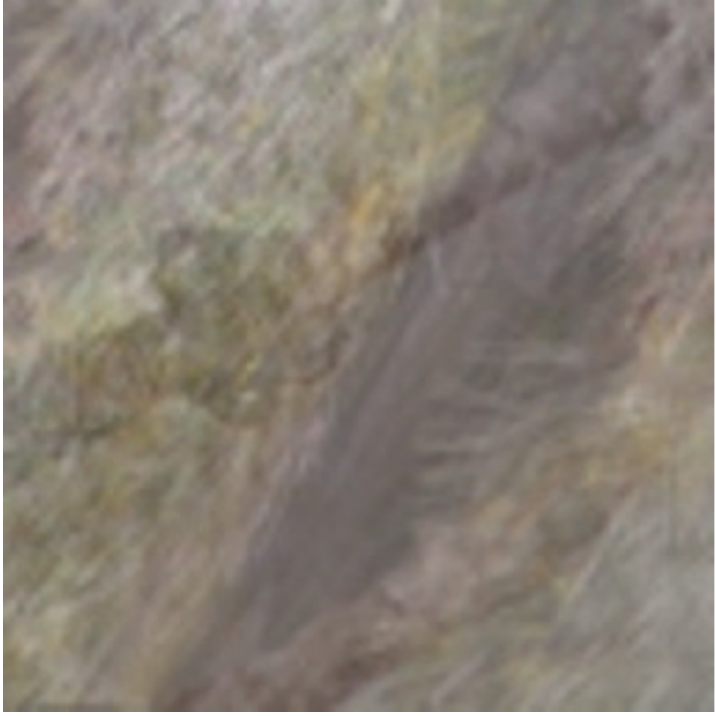}
\end{minipage}%
}}

\caption{Pipeline of deformable part descriptor (DPD) on a sample test images. It uses DPM for part localization and then use learned pooling weights for final pose-normalized representation.}
\label{fig:birds}
\end{figure}

Our second approach, we tested DeCAF in a pose-normalized setting using the deformable part descriptors (DPD) method~\cite{dpd}. Inspired by the deformable parts model \cite{dpm}, DPD explicitly utilizes the part localization to do semantic pooling. Specifically, after training a weakly-supervised DPM on bird images, the pool weight for each part of each component is calculated by using the key-point annotations to get cross-component semantic part correspondence. The final pose-normalized representation is computed by pooling the image features of predicted part boxes using the pooling weights. Based on the DPD implementation provided by the authors, we applied DeCAF in the same pre-trained DPM model and part predictions and used the same pooling weights. Figure \ref{fig:birds} shows the DPM detections and visualization of pooled DPD features on a sample test image.  As our first approach, we resized each predicted part box to 256 $\times$ 256 and computed DeCAF$_6$ to replace the KDES image features \cite{kdes} used in DPD paper.

Our performance as well as those from the literature are listed in Table \ref{table:birds}. DeCAF together with a simple logistic regression already obtains a significant performance increase over existing approaches, indicating that such features, although not specifically designed to model subcategory-level differences, captures such information well. In addition, explicitly taking more structured information such as part locations still helps, and provides another significant performance increase, obtaining an accuracy of 64.96\%, compared to the 50.98\% accuracy reported in~\cite{dpd}. It also outperforms POOF \cite{poof}, which is the best part-based approach for fine-grained categorization published so far. 

To the best of our knowledge, this is the best accuracy reported so far in the literature.

\begin{table}
\centering
\begin{tabular}{lc}
\toprule
Method & Accuracy\\
\midrule
DeCAF$_6$  & 58.75 \\
DPD + DeCAF$_6$ & \bf{64.96} \\
\\
DPD \cite{dpd}& 50.98 \\
POOF \cite{poof}& 56.78 \\
\bottomrule
\end{tabular}
\caption{Accuracy on the Caltech-UCSD bird dataset.}
\label{table:birds}
\end{table}

We note again that in all the experiments above, no fine-tuning is carried out on the CNN layers
since our main interest is to analyze how DeCAF generalizes to different tasks.
To obtain the best possible result one may want to perform a full back-propagation. However, the fact
that we see a significant performance increase without fine-tuning suggests that
DeCAF may serve as a good off-the-shelf visual representation without heavy computation.

\subsection{Scene recognition}
\label{sec:sun}

Finally, we evaluate DeCAF on the SUN-397 large-scale scene recognition database~\citep{xiao10}.
Unlike object recognition, wherein the goal is to identify and classify an object which is usually the primary focus of the image, the goal of a scene recognition task is to classify the \textit{scene} of the entire image.
In the SUN-397 database, there are 397 semantic scene categories including \textit{abbey}, \textit{diner}, \textit{mosque}, and \textit{stadium}.
Because DeCAF is learned on ILSVRC, an object recognition database, we are applying it to a task for which it was not designed.
Hence we might expect this task to be very challenging for these features, unless they are highly generic representations of the visual world.

Based on the success of using dropout with DeCAF$_6$ and DeCAF$_7$ for the object recognition task detailed in Section~\ref{sec:caltech}, we train and evaluate linear classifiers on these dropped-out features on the SUN-397 database.
Table~\ref{tab:sunresults} gives the classification accuracy results averaged across 5 splits of 50 training images and 50 test images.
Parameters are fixed for all methods, but we select the top-performing method by cross-validation, training on 42 images and testing on the remaining 8 in each split.

Our top-performing method in terms of cross-validation accuracy was to use DeCAF$_7$ with the SVM classifier, resulting in 40.94\% test performance.
Comparing against the method of~\citet{xiao10}, the current state-of-the-art method, we see a performance improvement of 2.9\% using only DeCAF.
Note that, like the state-of-the-art method used as a baseline in Section~\ref{sec:caltech}, this method uses a large set of traditional vision features and combines them with a multi-kernel learning method.
The fact that a simple linear classifier on top of our single image feature outperforms the multi-kernel learning baseline built on top of many traditional features demonstrates the ability of DeCAF to generalize to other tasks and its representational power as compared to traditional hand-engineered features.

\begin{table}
\centering

\begin{tabular}{lcc}
\toprule
& DeCAF$_6$ & DeCAF$_7$ \\
\midrule
LogReg & $\bm{ 40.94 \pm 0.3}$ & $40.84 \pm 0.3$ \\
SVM & $39.36 \pm 0.3$ & $40.66 \pm 0.3$ \\
\\
\citet{xiao10} & \multicolumn{2}{c}{38.0} \\
\bottomrule
\end{tabular}
\caption{Average accuracy per class on SUN-397 with 50 training samples and 50 test samples per class, across two hidden layers of the network and two classifiers.
Our result from the training protocol/classifier combination with the best validation accuracy -- Logistic Regression with DeCAF$_7$ -- is shown in \textbf{bold}.
\label{tab:sunresults}
}
\vspace{-.5cm}
\end{table}

\section{Discussion}
In this work, we analyze the use of deep features applied in a semi-supervised multi-task framework.
In particular, we demonstrate that by leveraging an auxiliary large labeled object database to train a deep convolutional architecture, we can learn features that have sufficient representational power and generalization ability to perform semantic visual discrimination tasks using simple linear classifiers, reliably outperforming current state-of-the-art approaches based on sophisticated multi-kernel learning techniques with traditional hand-engineered features.
Our visual results demonstrate the generality and semantic knowledge implicit in these features, showing that the features tend to cluster images into interesting semantic categories on which the network was never explicitly trained.
Our numerical results consistently and robustly demonstrate that our multi-task feature learning framework can substantially improve the performance of a wide variety of existing methods across a spectrum of visual recognition tasks, including domain adaptation, fine-grained part-based recognition, and large-scale scene recognition.
The ability of a visual recognition system to achieve high classification accuracy on tasks with sparse labeled data has proven to be an elusive goal in computer vision research, but our multi-task deep learning framework and fast open-source implementation are significant steps in this direction.
While our current experiments focus on contemporary recognition challenges, we expect our feature to be very useful in detection, retrieval, and category discovery settings as well.

\section{Acknowledgements}

The authors would like to thank Alex Krizhevsky for his valuable help in reproducing the ILSVRC-2012 results, as well as for providing an open source implementation of GPU-based CNN training. The authors also thank NSF, DoD, Toyota, and the Berkeley Vision and Learning Center sponsors for their support to our research group.

\footnotesize{

}
\end{document}